\documentclass[runningheads]{llncs}
\usepackage[T1]{fontenc}
\usepackage{graphicx}
\usepackage{orcidlink}
\usepackage{booktabs}
\usepackage{multirow}
\usepackage{xspace}
\usepackage{booktabs}
\usepackage{amsmath}
\usepackage{pifont}
\usepackage{xcolor}
\usepackage{url}

\def \ours {CarPatch\xspace}
\def \oursnospace {CarPatch}
\def \etal {\emph{et al.}}
\def \ie {\emph{i.e.}}
\def \eg {\emph{e.g.}}

\newcommand{\tit}[1]{\smallskip\noindent\textbf{#1.}}
\newcommand{\cmark}{\ding{51}}
\newcommand{\xmark}{\ding{55}}

\begin{document}
\title{\textit{\ours}: A Synthetic Benchmark for Radiance Field Evaluation on Vehicle Components
}
\titlerunning{\textit{\ours}}
%
\author{
 Davide Di Nucci \orcidlink{0009-0000-7450-8796} \inst{1} \and
 Alessandro Simoni \orcidlink{0000-0003-3095-3294} \inst{1} \and 
 Matteo Tomei \orcidlink{0000-0002-1385-924X} \inst{2} \and
 Luca Ciuffreda\inst{2} \and \\
 Roberto Vezzani \orcidlink{0000-0002-1046-6870} \inst{1} \and
 Rita Cucchiara \orcidlink{0000-0002-2239-283X} \inst{1}
}
\authorrunning{D. Di Nucci \etal}
%
\institute{
 Department of Engineering “Enzo Ferrari” (DIEF), University of Modena and
 Reggio Emilia, 41125 Modena, Italy
 \email{\{davide.dinucci,alessandro.simoni,roberto.vezzani,rita.cucchiara\}@unimore.it}\\
 \and
 Prometeia, 40137 Bologna, Italy \\
 \email{\{matteo.tomei,luca.ciuffreda\}@prometeia.com}
}
\maketitle              

\begin{figure}[htb]
\vspace{15pt}
\begin{center}
\includegraphics[width=\linewidth]{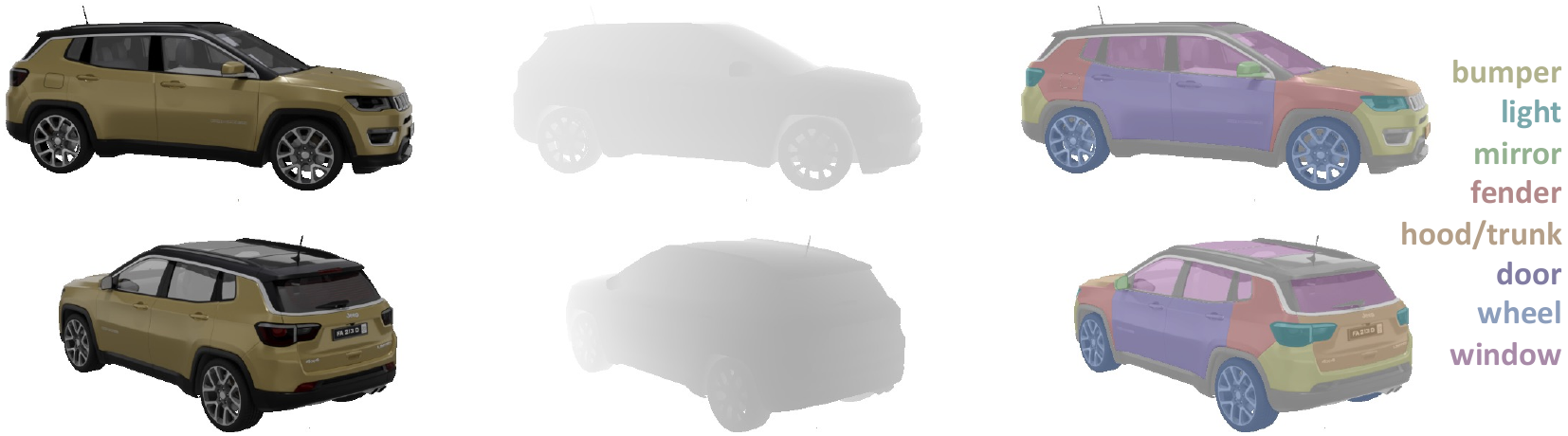}
\end{center}
\vspace{-0.25cm}
\caption{A visualization of the \textit{\ours} data: RGB images (left), depth images (center), and semantic segmentation of vehicle components (right).}
\label{fig:first_page}
\vspace{0.5cm}
\end{figure}

\vspace{-30pt}
\begin{abstract}

Neural Radiance Fields (NeRFs) have gained widespread recognition as a highly effective technique for representing 3D reconstructions of objects and scenes derived from sets of images. Despite their efficiency, NeRF models can pose challenges in certain scenarios such as vehicle inspection, where the lack of sufficient data or the presence of challenging elements (\eg~reflections) strongly impact the accuracy of the reconstruction. To this aim, we introduce \textit{\ours}, a novel synthetic benchmark of vehicles. In addition to a set of images annotated with their intrinsic and extrinsic camera parameters, the corresponding depth maps and semantic segmentation masks have been generated for each view. 
Global and part-based metrics have been defined and used to evaluate, compare, and better characterize some state-of-the-art techniques. 
The dataset is publicly released at \url{https://aimagelab.ing.unimore.it/go/carpatch} and can be used as an evaluation guide and as a baseline for future work on this challenging topic.

\keywords{Synthetic vehicle dataset \and 3D Reconstruction \and Neural radiance fields \and Volumetric rendering \and RGB-D.}
\end{abstract}

\setcounter{footnote}{0}

\section{Introduction}
\label{sec:introduction}

Recent advances in Neural Radiance Fields (NeRFs)~\cite{mildenhall2021nerf} strongly improved the fidelity of generated novel views by fitting a neural network to predict the volume density and the emitted radiance of each 3D point in a scene. The differentiable volume rendering step allows having a set of images, with known camera poses, as the only input for model fitting. Moreover, the limited amount of data, \ie~(image, camera pose) pairs, needed to train a NeRF model, facilitates its adoption and drives the increasing range of its possible applications. Among these, view synthesis recently emerged for street view reconstruction~\cite{muller2022autorf,xie2023snerf} in the context of AR/VR applications, robotics, and autonomous driving, with considerable efforts towards vehicle novel view generation. However, these attempts focus on images representing large-scale unbounded scenes, such as those from KITTI~\cite{geiger2012we}, and usually fail to achieve high-quality 3D vehicle reconstruction.

In this paper, we introduce an additional use case for neural radiance fields, \ie~\textit{vehicle inspection}, where the goal is to represent an individual high-quality instance of a given car. The availability of a high-fidelity 3D vehicle representation could be beneficial whenever the car body has to be analyzed in detail. For instance, insurance companies or body shops could rely on NeRF-generated views to assess possible external damages after a road accident and estimate their repair cost. Moreover, rental companies could compare two NeRF models trained before and after each rental, respectively, to assign responsibility for any new damages. This would avoid expert on-site inspection or a rough evaluation based on a limited number of captures.

For this purpose, we provide an experimental overview of the state-of-the-art NeRF methods, suitable for vehicle reconstruction. To make the experimental setting reproducible and to provide a basis for new experimentation, we propose \textit{\ours}, a new benchmark to assess neural radiance field methods on the~\textit{vehicle inspection} task. Specifically, we generate a novel dataset consisting of 8 different synthetic scenes, corresponding to as many high-quality 3D car meshes with realistic details and challenging light conditions. As depicted in Fig.~\ref{fig:first_page}, we provide not only RGB images with camera poses, but also binary masks of different car components to validate the reconstruction quality of specific vehicle parts (\eg~wheels or windows). Moreover, for each camera position, we generate the ground truth depth map with the double goal of examining the ability of NeRF architectures to correctly predict volume density and, at the same time, enable future works based on RGB-D inputs. We evaluate the novel view generation and depth estimation performance of several methods under diverse settings (both global and component-level). Finally, since the process of image collection for fitting neural radiance fields could be time consuming in real scenarios, we provide the same scenes by varying the number of training images, in order to determine the robustness to the amount of training data.

After an overview of the main related works in Sec.~\ref{sec:related}, we thoroughly describe the process of 3D mesh gathering, scene setup, and dataset generation in Sec.~\ref{sec:dataset}. The evaluation of existing NeRF architectures on \textit{\ours} is presented in Sec.~\ref{sec:benchmark}.

\section{Related work}
\label{sec:related}
We provide a brief overview of the latest updates in neural radiance field, including its significant extensions and applications that have influenced our work.
NeRF limitations have been tackled by different works, trying to reduce its complexity, increase the reconstruction quality, and develop more challenging benchmarks.

\tit{Neural scene reconstruction}
The handling of aliasing artifacts is a well-known issue in rendering algorithms. Mip-NeRF~\cite{barron2021mip,barron2022mip} and Zip-NeRF~\cite{barron2023zip} have tackled the aliasing issue by reasoning on volumetric frustums along a cone. These approaches have inspired works such as Able-NeRF~\cite{tang2023able}, which replaces the MLP of the original implementation with a transformer-based architecture.
In addition to other sources of aliasing, reflections can pose a challenge for NeRF. 
Several works have attempted to address the issue of aliasing in reflections by taking into account the reflectance of the scene ~\cite{bi2020neural,boss2021neural,verbin2022ref}. 
Moreover, computation is a widely recognized concern. Various works in the literature have demonstrated that it is possible to achieve high-fidelity reconstructions while reducing the overall training time. 
Two notable works in this direction include 
NSVF~\cite{liu2020neural}, which uses a voxel-based representation for more efficient rendering of large scenes, and
Instant-NGP~\cite{muller2022instant}, which proposes a multi-resolution hash table combined with a light MLP to achieve faster training times. Other approaches such as DVGO~\cite{sun2022direct} and Plenoxels~\cite{yu2021plenoxels} optimize voxel grids of features to enable fast radiance field reconstruction. TensoRF~\cite{chen2022tensorf} combines the traditional CP decomposition [7] with a new vector-matrix decomposition method~\cite{carroll1970analysis} leading to faster training and higher-quality reconstruction. 

In this work, in order to satisfy real-time performances for vehicle inspection, we select a set of architectures that strike a balance between training time and the quality of the reconstruction.

\begin{table*}[t]
    \caption{Comparison between existing datasets used as benchmarks for neural radiance field evaluation and \textit{\ours}. We provide the same scene by varying the amount of training data (40, 60, 80, and 100 images), allowing users to test the robustness of their architectures. We also release depth and segmentation data for all the images.}
    \centering
    \begin{tabular}{l c c c c }
        \toprule
        Dataset & Scenes & Images/scene & Depth & Segmentation\\
        \midrule
        Blender~\cite{mildenhall2021nerf} & 8 & 300 & \cmark & \xmark \\
        Shiny Blender~\cite{verbin2022ref} & 6 & 300 & \cmark & \xmark \\
        BlendedMVG~\cite{yao2020blendedmvs} & 508 & 200-4000 & \xmark & \xmark \\
        \midrule
        \textit{\oursnospace}$_{40}$  & 8 & 240 & \cmark & \cmark \\
        \textit{\oursnospace}$_{60}$  & 8 & 260 & \cmark & \cmark \\
        \textit{\oursnospace}$_{80}$  & 8 & 280 & \cmark & \cmark \\
        \textit{\oursnospace}$_{100}$  & 8 & 300 & \cmark & \cmark \\

        \bottomrule
    \end{tabular}
    \label{tab:dataset_comparison}
\end{table*}

\tit{Scene representation benchmarks}
One of the most widely used benchmarks for evaluating NeRF is the Nerf Synthetic Blender dataset~\cite{mildenhall2021nerf}. This dataset consists of 8 different scenes generated using Blender\footnote{\url{http://www.blender.org}}, each with 100 training images and 200 test images. Other synthetic datasets include the Shiny Blender dataset~\cite{verbin2022ref}, which mostly contains singular objects with simple geometries, and Blend DMVS \cite{yao2020blendedmvs}, which provides various scenes to test NeRF implementations at different scales.
These works do not provide ground truth information about the semantic meaning of the images. This limitation makes it difficult to study the ability of NeRF to reconstruct certain surfaces compared to others. In our \textit{\ours} dataset, we provide ground truth segmentation of vehicle components in the scene, allowing for the evaluation of architectures on specific parts. Table~\ref{tab:dataset_comparison} presents a comparison between the most common datasets used as benchmarks and our proposed dataset.


\section{The \textit{\ours} dataset}

\begin{figure*}[t]
    \centering
    \includegraphics[width=0.98\linewidth]{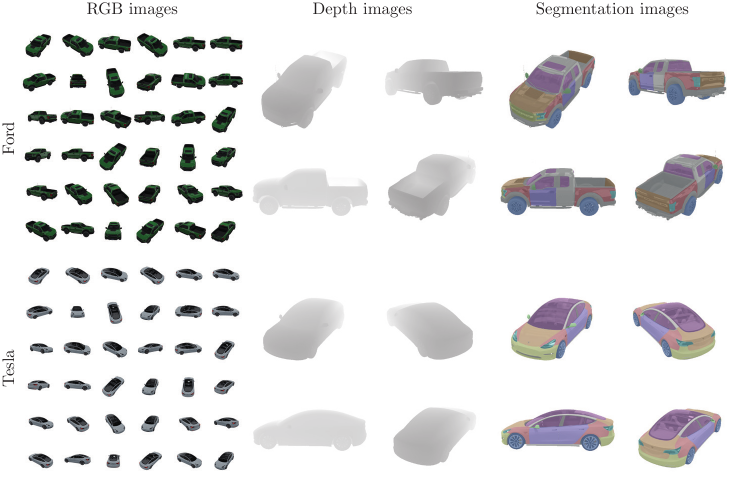}
    \caption{Sample RGB images (left), depth data (center), and segmentation masks (right) from \textit{\ours}, for different car models.}
    \label{fig:dataset}
\end{figure*}

\label{sec:dataset}
In this section, we detail the source data and the procedure exploited for generating our \textit{\ours} dataset. In particular, we describe how we gathered 3D models, set up the blender scenes, and designed the image capture process.

\subsection{Synthetic 3D models and scene setup}
All the 3D models included in \textit{\ours} scenes have been downloaded from Sketchfab\footnote{\url{https://sketchfab.com}}, a large collection of free 3D objects for research use.
Table \ref{tab:3d_models} provides a detailed list of all the starting models used. Each of them has been edited in Blender to enhance its realism; specifically, we improved the materials, colors, and lighting in each scene to create a more challenging environment.

The scenes have been set up accordingly to the Google Blender dataset~\cite{mildenhall2021nerf}. The lighting conditions and rendering settings were customized to create a more realistic environment. The vehicle was placed at the center of the scene at position (0,0,0), with nine lights distributed around the car and varying emission strengths to create shadows and enhance reflections on the materials' surfaces. To improve realism, we resized objects to match their real-world size. The camera and lights were placed in order to provide an accurate representation of the environment, making the scenes similar to real-world scenarios.

\begin{table*}[t]
    \centering
    \caption{Summary of the source 3D models from which our dataset has been generated, including their key features.}
    \begin{tabular}{l c c c c c }
        \toprule
        Model name & Acronym & \#Triangles & \#Vertices & \#Textures & \#Materials\\
        \midrule
        Tesla Model & \textsc{Tesla} & 684.3k & 364.4k & 22 & 58 \\
        Smart & \textsc{Smart} & 42.8k & 26.4k & 0 & 31 \\
        Ford Raptor & \textsc{Ford} & 257.1k & 156.5k & 12 & 50 \\
        BMW M3 E46 & \textsc{Bmw} & 846.9k & 442.4k & 7 & 39 \\
        Mercedes GLK & \textsc{Mbz$_1$} & 1.3M &  741.4k & 0 & 15 \\
        Mercedes CLS & \textsc{Mbz$_2$} & 1.0M & 667k & 0 & 18 \\
        Volvo S90 & \textsc{Volvo} & 3.3M & 1.7M & 56 & 44 \\
        Jeep Compass & \textsc{Jeep} & 334.7k & 189.6k & 7 & 39 \\
        \bottomrule
    \end{tabular}
    \label{tab:3d_models}
\end{table*}

\subsection{Dataset building}
The dataset was built using the Python interface provided in Blender, allowing us to control objects in the environment. For each rendered image, we captured not only the RGB color values but also the corresponding depth map, as well as the pixel-wise semantic segmentation masks for eight vehicle components: bumpers, lights, mirrors, hoods/trunks, fenders, doors, wheels, and windows. Examples of these segmentation masks can be seen in Fig.~\ref{fig:first_page}. Please note that all the pixels belonging to a component (\eg~doors) are grouped into the same class, regardless of the specific component location (\eg~front/rear/right/left door). The bpycv\footnote{\url{https://github.com/DIYer22/bpycv}} utility has been used for collecting additional metadata, enabling us to evaluate NeRF models on the RGB reconstruction and depth estimation of the overall vehicle as well as each of its subparts.

For the rendering of training images, the camera randomly moved on the hemisphere centered in (0,0,0) and above the ground. The camera rotation angle was sampled from a uniform distribution before each new capture. For building the test set, the position of the camera was kept at a fixed distance from the ground and rotated around the Z-axis with a fixed angle equal to $\frac{2\pi}{\#test\_views}$ radians before each new capture.

In order to guarantee the fairness of the current and future comparisons, we explicitly provide four different versions of each scene, by varying the number of training images (40, 60, 80, and 100 images, respectively). Different versions of the same scene have no overlap in training camera poses, while the test set is always the same and contains 200 images for each scene.\\ We release the code for dataset creation and metrics evaluation at \url{https://github.com/davidedinuc/carpatch}.

\section{Benchmark}
\label{sec:benchmark}

This section presents the selection and testing of various recent NeRF-based methods~\cite{muller2022instant,chen2022tensorf,sun2022direct} on the presented \textit{\ours} dataset, with a detailed description of the experimental setting for each baseline. Additionally, we assess the quality of the reconstructed vehicles in terms of their appearance and 3D surface reconstruction, utilizing depth maps generated during volume rendering.

\subsection{Compared methods}
To overcome challenges related to illumination and reflective surfaces during the process of reconstructing vehicles, it is crucial to choose an appropriate neural rendering approach.
We tested selected approaches on \textit{\ours} without modifying the implementation details available in the original repositories, whenever possible. However, some parameters had to be adjusted in order to fit our models (which are larger compared to reference dataset meshes) to the scene. All tests were performed on a GeForce GTX 1080 Ti. 
After considering various NeRF systems, we have selected the following baselines:
\begin{itemize}
 \item \tit{Instant-NGP~\cite{muller2022instant}} Since the original implementation of Instant-NGP is in CUDA, we decided to use an available PyTorch implementation\footnote{\url{https://github.com/kwea123/ngp\_pl}} of this approach in order to have a fair comparison with the other approaches. In our experiments, a batch size of 8192 was maintained, with a scene scale of 0.5 and a total of 30,000 iteration steps.
 \item \tit{TensoRF~\cite{chen2022tensorf}} In our setting, a batch of 4096 rays was used. Additionally, we increased the overall scale of the scene from 1 to 3.5. These adjustments were made after experimentation and careful consideration of the resulting reconstructions. Training lasts 30,000 iterations.
\item \tit{DVGO~\cite{sun2022direct}} In this work, the training process consists of two phases: a coarse training phase of 5,000 iterations, followed by a fine training phase of 20,000 iterations that aims to improve the model's ability to learn intricate details of the scene. In our experiments, we applied a batch size of 8192 while maintaining the default scene size.
\end{itemize}

\begin{table*}[t]
    \centering
    \caption{Quantitative results on the \textit{\ours} test set for each vehicle model.}
    \begin{tabular}{l c c c c c c c c c c}
        \toprule
        Method & Metric & \textsc{Bmw} & \textsc{Tesla} & \textsc{Smart} & \textsc{Mbz$_1$} & \textsc{Mbz$_2$} & \textsc{Ford} & \textsc{Jeep} & \textsc{Volvo} & \textit{Avg} \\
        \midrule
        iNGP~\cite{muller2022instant} & \multirow{3}{*}{PSNR$\uparrow$} & $39.48$ & $39.46$ & $39.57$ & $36.87$ & $39.15$ & $33.67$ & $35.00$ & $35.93$ & $37.39$ \\
        DVGO~\cite{sun2022direct} & & $39.91$ & $39.89$ & $40.34$ & $37.45$ & $39.37$ & $33.82$ & $35.32$ & $36.28$ & $37.80$ \\
        TensoRF~\cite{chen2022tensorf} & & $40.68$ & $39.92$ & $40.38$ & $38.07$ & $40.84$ & $34.33$ & $34.87$ & $36.77$ & $\mathbf{38.23}$ \\
        \midrule
        iNGP~\cite{muller2022instant} & \multirow{3}{*}{SSIM$\uparrow$} & $0.985$ & $0.987$ & $0.988$ & $0.985$ & $0.987$ & $0.959$ & $0.978$ & $0.979$ & $0.981$ \\
        DVGO~\cite{sun2022direct} & & $0.987$ & $0.988$ & $0.990$ & $0.987$ & $0.988$ & $0.964$ & $0.980$ & $0.981$ & $0.983$ \\
        TensoRF~\cite{chen2022tensorf} & & $0.989$ & $0.987$ & $0.99$ & $0.989$ & $0.991$ & $0.966$ & $0.975$ & $0.982$ & $\mathbf{0.984}$ \\
        \midrule
        iNGP~\cite{muller2022instant} & \multirow{3}{*}{LPIPS$\downarrow$} & $0.029$ & $0.029$ & $0.02$ & $0.028$ & $0.024$ & $0.062$ & $0.036$ & $0.032$ & $0.032$ \\
        DVGO~\cite{sun2022direct} & & $0.022$ & $0.022$ & $0.014$ & $0.019$ & $0.020$ & $0.051$ & $0.029$ & $0.022$ & $\mathbf{0.025}$ \\
        TensoRF~\cite{chen2022tensorf} & & $0.023$ & $0.026$ & $0.017$ & $0.02$ & $0.017$ & $0.051$ & $0.039$ & $0.027$ & $0.028$ \\
        \midrule
        iNGP~\cite{muller2022instant} & \multirow{3}{*}{D-RMSE$\downarrow$} & $0.640$ & $0.369$ & $0.377$ & $0.496$ & $0.500$ & $0.406$ & $0.558$ & $0.674$ & $0.503$ \\
        DVGO~\cite{sun2022direct} & & $0.561$ & $0.353$ & $0.305$ & $0.437$ & $0.454$ & $0.339$ & $0.469$ & $0.561$ & $\mathbf{0.435}$ \\
        TensoRF~\cite{chen2022tensorf} & & $0.590$ & $0.357$ & $0.335$ & $0.467$ & $0.482$ & $0.375$ & $0.536$ & $0.626$ & $0.471$ \\
        \midrule
        iNGP~\cite{muller2022instant} & \multirow{3}{*}{SN-RMSE$\downarrow$} & $4.24$ & $3.38$ & $3.41$ & $4.26$ & $4.13$ & $5.15$ & $4.60$ & $4.67$ & $4.23$ \\
        DVGO~\cite{sun2022direct} & & $4.27$ & $3.48$ & $3.20$ & $4.19$ & $4.24$ & $5.04$ & $4.67$ & $4.71$ & $4.22$ \\
        TensoRF~\cite{chen2022tensorf} & & $3.96$ & $3.24$ & $3.10$ & $4.00$ & $3.91$ & $4.91$ & $4.41$ & $4.48$ & $\mathbf{4.00}$ \\
        \bottomrule
    \end{tabular}
    \label{tab:category-results}
\end{table*}

\begin{table*}[t]
    \centering
    \caption{Quantitative results on the \textit{\ours} test set for each vehicle component averaged over the vehicle models.}
    \begin{tabular}{l c c c c c c }
        \toprule
        Method & Component & PSNR$\uparrow$ & SSIM$\uparrow$ & LPIPS$\downarrow$ & D-RMSE$\downarrow$ & SN-RMSE$\downarrow$ \\
        \midrule
        iNGP~\cite{muller2022instant} & \multirow{3}{*}{\textit{bumper}} & $33.05$ & $0.986$ & $0.019$ & $0.281$ & $0.79$ \\
        DVGO~\cite{sun2022direct} & & $34.41$ & $0.989$ & $0.011$ & $\mathbf{0.236}$ & $0.72$ \\
        TensoRF~\cite{chen2022tensorf} & & $\mathbf{35.49}$ & $\mathbf{0.991}$ & $\mathbf{0.010}$ & $0.311$ & $\mathbf{0.68}$ \\
        \midrule
        iNGP~\cite{muller2022instant} & \multirow{3}{*}{\textit{light}} & $28.71$ & $0.993$ & $0.009$ & $0.421$ & $0.48$ \\
        DVGO~\cite{sun2022direct} & & $29.10$ & $0.995$ & $\mathbf{0.006}$ & $\mathbf{0.384}$ & $0.43$ \\
        TensoRF~\cite{chen2022tensorf} & & $\mathbf{29.68}$ & $\mathbf{0.996}$ & $\mathbf{0.006}$ & $0.438$ & $\mathbf{0.38}$ \\
        \midrule
        iNGP~\cite{muller2022instant} & \multirow{3}{*}{\textit{mirror}} & $29.60$ & $0.994$ & $0.011$ & $0.427$ & $0.43$ \\
        DVGO~\cite{sun2022direct} & & $31.16$ & $\mathbf{0.996}$ & $\mathbf{0.007}$ & $\mathbf{0.345}$ & $\mathbf{0.38}$ \\
        TensoRF~\cite{chen2022tensorf} & & $\mathbf{31.68}$ & $\mathbf{0.996}$ & $0.008$ & $0.372$ & $0.39$ \\
        \midrule
        iNGP~\cite{muller2022instant} & \multirow{3}{*}{\textit{hood/trunk}} & $32.28$ & $0.977$ & $0.052$ & $0.260$ & $1.33$ \\
        DVGO~\cite{sun2022direct} & & $32.68$ & $0.981$ & $\mathbf{0.038}$ & $\mathbf{0.259}$ & $1.35$ \\
        TensoRF~\cite{chen2022tensorf} & & $\mathbf{33.75}$ & $\mathbf{0.983}$ & $0.040$ & $0.302$ & $\mathbf{1.24}$ \\
        \midrule
        iNGP~\cite{muller2022instant} & \multirow{3}{*}{\textit{fender}} & $32.44$ & $0.990$ & $0.021$ & $0.253$ & $0.87$ \\
        DVGO~\cite{sun2022direct} & & $33.55$ & $\mathbf{0.993}$ & $\mathbf{0.013}$ & $\mathbf{0.223}$ & $0.85$ \\
        TensoRF~\cite{chen2022tensorf} & & $\mathbf{34.36}$ & $\mathbf{0.993}$ & $0.015$ & $0.267$ & $\mathbf{0.77}$ \\
        \midrule
        iNGP~\cite{muller2022instant} & \multirow{3}{*}{\textit{door}} & $34.19$ & $0.969$ & $0.079$ & $0.182$ & $0.67$ \\
        DVGO~\cite{sun2022direct} & & $35.48$ & $0.977$ & $\mathbf{0.042}$ & $\mathbf{0.173}$ & $0.74$ \\
        TensoRF~\cite{chen2022tensorf} & & $\mathbf{36.25}$ & $\mathbf{0.979}$ & $0.051$ & $0.191$ & $\mathbf{0.62}$ \\
        \midrule
        iNGP~\cite{muller2022instant} & \multirow{3}{*}{\textit{wheel}} & $33.12$ & $0.995$ & $0.008$ & $0.391$ & $0.87$ \\
        DVGO~\cite{sun2022direct} & & $33.65$ & $0.995$ & $0.006$ & $\mathbf{0.267}$ & $\mathbf{0.79}$ \\
        TensoRF~\cite{chen2022tensorf} & & $\mathbf{34.55}$ & $\mathbf{0.996}$ & $\mathbf{0.005}$ & $0.334$ & $\mathbf{0.79}$ \\
        \midrule
        iNGP~\cite{muller2022instant} & \multirow{3}{*}{\textit{window}} & $26.44$ & $0.897$ & $0.166$ & $0.879$ & $2.52$ \\
        DVGO~\cite{sun2022direct} & & $26.54$ & $\mathbf{0.899}$ & $\mathbf{0.147}$ & $\mathbf{0.779}$ & $2.57$ \\
        TensoRF~\cite{chen2022tensorf} & & $\mathbf{26.74}$ & $0.896$ & $0.160$ & $0.834$ & $\mathbf{2.38}$ \\
        \bottomrule
    \end{tabular}
    \label{tab:panels-results}
\end{table*}

\subsection{Metrics}
\label{sec:metrics}
The effectiveness of the chosen methods has been assessed thanks to the typical perceptual metrics used in NeRF-based reconstruction tasks, namely PSNR, SSIM~\cite{wang2004image}, and LPIPS~\cite{zhang2018unreasonable}. 

However, the appearance-based metrics are strongly related to the emitted radiance besides the learned volume density. We suggest two supplementary depth-based metrics for the sole purpose of assessing the volume density. Since it is not feasible to obtain ground truth 3D models of the vehicles in real-world scenarios, we utilize the depth map as our knowledge of the 3D surface of the objects. Specifically, we define a depth map as a matrix
\begin{equation}
    D = \{d_{ij}\}, d_{ij} \in [0, R]
\end{equation}
in which each value $d_{ij}$ ranges from $0$ to the maximum depth value $R$. Furthermore, we estimate the surface normals from the depth maps~\cite{Pini21Depth}. Initially, we establish the orientation of a surface normal as:
\begin{equation}
    \mathbf{d} = \langle d_x, d_y, d_z \rangle = \left(-\frac{\partial d_{ij}}{\partial i}, -\frac{\partial d_{ij}}{\partial j}, 1 \right) \approx \left( d_{(i+1)j} - d_{ij}, d_{i(j+1)} - d_{ij}, 1 \right)
\end{equation}
where the first two elements represent the depth gradients in the $i$ and $j$ directions, respectively. Afterward, we normalize the normal vector to obtain a unit-length vector $\mathbf{n}(d_{ij}) = \frac{\mathbf{d}}{\left\lVert \mathbf{d} \right\rVert}$.

We assess the 3D reconstruction's quality through the following metrics:
\begin{itemize}
    \item \tit{Depth Root Mean Squared Error (D-RMSE)} This metric measures the average difference in meters between the ground truth and predicted depth maps.
    \begin{equation}
        \textnormal{D-RMSE} = \sqrt{\frac{\sum_{i=0}^{M}\sum_{j=0}^{N} (\hat{d}_{ij} - d_{ij})^2}{M \cdot N}}
    \end{equation}
    \item \tit{Surface Normal Root Mean Squared Error (SN-RMSE)} This metric measures the average angular error in degrees between the angle direction of the ground truth and predicted surface normals.
    \begin{equation}
        \textnormal{SN-RMSE} = 
        \sqrt{\frac{\sum_{i=0}^{M}\sum_{j=0}^{N} (\arccos(\mathbf{n}(\hat{d}_{ij})) - \arccos(\mathbf{n}(d_{ij})))^2}{M \cdot N}}
    \end{equation}
\end{itemize}
D-RMSE and SN-RMSE are computed only for those pixels with a positive depth value in both GT and predicted depth maps. This avoids computing depth estimation errors on background pixels (which have a fixed depth value of 0).

\begin{figure}[t]
    \centering
    \includegraphics[width=0.99\linewidth]{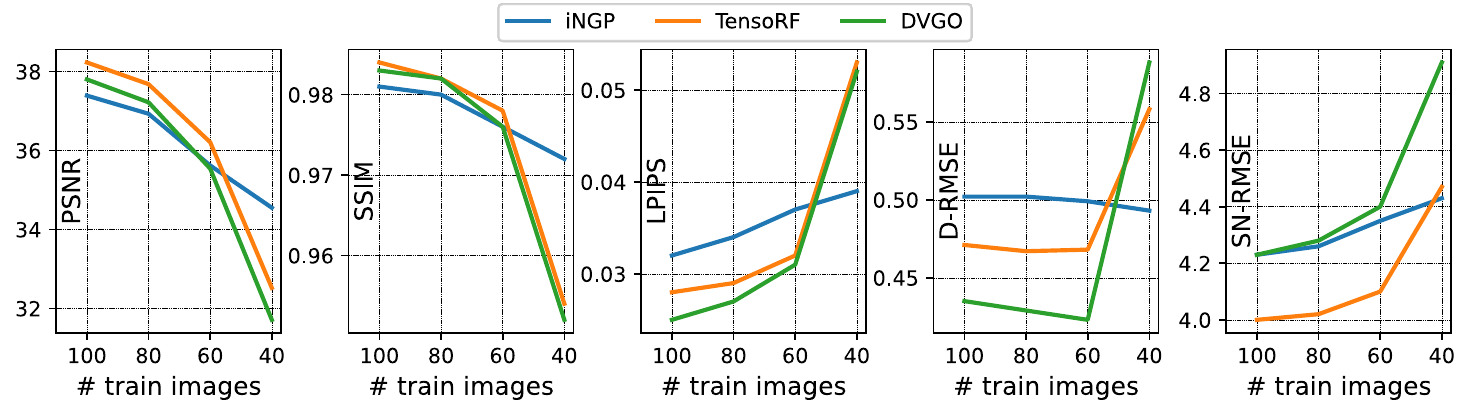}
    \caption{Performance by varying the number of training images, in terms of PSNR, SSIM, LPIPS, D-RMSE, and SN-RMSE. Despite its lower overall performance, Instant-NGP~\cite{muller2022instant} exhibits low variance with respect to the amount of training data.}
    \label{fig:metrics_num_imgs}
\end{figure}

\begin{figure}[t]
    \centering
    \includegraphics[width=0.98\linewidth]{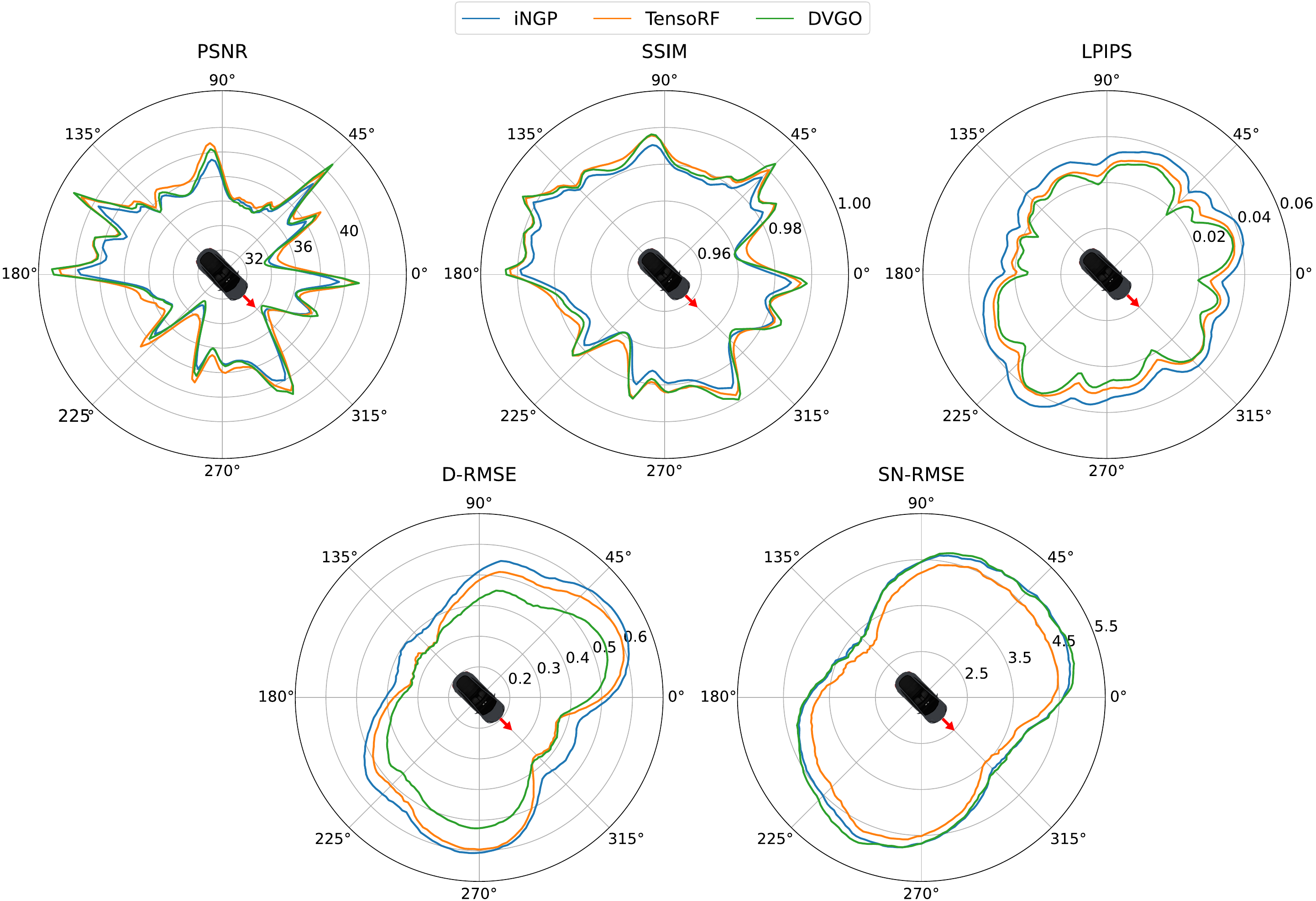}
    \caption{Performance by camera viewing angle, in terms of PSNR, SSIM, LPIPS, D-RMSE, and SN-RMSE. Depending on the training camera distribution, all the methods struggle wherever the viewpoints are more sparse (\eg~between $225^\circ$ and $270^\circ$). The red arrow represents where the front of the vehicle is facing.}
    \label{fig:polar_plot}
\end{figure}

\subsection{Results}
The following section presents both quantitative and qualitative results obtained from the selected NeRF baselines. We will discuss their performance on the \textit{\ours} dataset, by analyzing the impact of viewing camera angle and the number of training images.

According to Table~\ref{tab:category-results}, all the selected NeRF approaches obtain satisfying results. Although the baselines demonstrate similar performances in terms of appearance scores (PSNR, SSIM, and LPIPS), our evaluation using depth-based metrics (D-RMSE and SN-RMSE) reveals significant differences in the 3D reconstruction of the vehicles. DVGO outperforms its competitors by achieving better depth estimation, resulting in a $+13.5\%$ improvement compared to iNGP and a $+7.6\%$ improvement compared to TensoRF. In contrast, TensoRF predicts a more accurate 3D surface with the lowest angular error on the surface normals.

\begin{figure*}[t]
    \centering
    \includegraphics[width=0.9\linewidth]{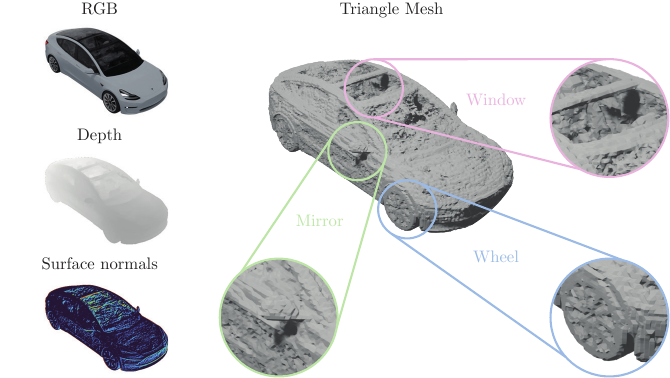}
    \caption{Sample of 3D reconstruction of the \textsc{Tesla}: (left) the reconstructed RGB, depth, and surface normals, (right) the reconstructed surfaces on the triangle mesh.}
    \label{fig:qualitatives}
\end{figure*}

Since our use case is related to vehicle inspection, in Table~\ref{tab:panels-results} we report results computed on each car component. For this purpose, we mask both GT and predictions using a specific component mask before computing the metrics. However, this would lead to an unbalanced ratio between background and foreground pixels, due to the limited components' area, and finally to a biased metric value. By computing D-RMSE and SN-RMSE only on foreground pixels (see Sec.~\ref{sec:metrics}), depth-based metrics are not affected by this issue. For PSNR, SSIM, and LPIPS, instead, we compute component-level metrics over the image crop delimited by the bounding boxes around each mask.
As expected, it is worth noting that NeRF struggles to reconstruct transparent objects (\eg~mirrors, lights, and windows) obtaining the highest errors in terms of depth and normal estimation. However, over the single components, TensoRF outperforms the competitors in most of the metrics and in particular on the surface normal estimation. The errors in the reconstruction of specific components' surfaces can also be appreciated in the qualitative results of Fig.~\ref{fig:qualitatives}.

Moreover, we analyze the performances of each method in terms of the number of training images. We trained the baselines on every version of the \textit{\ours} dataset and report the results in Fig.~\ref{fig:metrics_num_imgs}. It is worth noting that reducing the number of training images has a significant impact on all the metrics independently of the method. However, Instant-NGP demonstrates to be more robust to the number of camera viewpoints having a smoother drop, especially in terms of LPIPS, D-RMSE, and SN-RMSE.

Finally, we discuss how the training camera viewpoints' distribution around the vehicle may affect the performance of each method from certain camera angles. In particular, as depicted in Fig.~\ref{fig:polar_plot}, it is evident how between $180^\circ$ and $270^\circ$ and between $0^\circ$ and $45^\circ$ there are considerable variations in the metrics. Indeed, in these areas the datasets contain more sparsity in terms of camera viewpoints and, as expected, all the methods are affected.

\section{Conclusion}
\label{sec:conclusion}

In this article, we have proposed a new benchmark for the evaluation and comparison of NeRF-based techniques.
Focusing on one of the many concrete applications of this recent technology, i.e. \textit{vehicle inspection}, a new synthetic dataset including renderings of 8 vehicles was first created. 
In addition to the set of RGB views annotated with the camera pose, the dataset is enriched by semantic segmentation masks as well as depth maps to further analyze the results and compare the methods.

The presence of reflective surfaces and transparent parts makes the task of vehicle reconstruction still challenging.
Proposed additional metrics, as well as new graphical ways of displaying the results, are proposed to make these limitations more evident.
We are confident that \textit{\ours} can be of great help as a basis for research on NeRF models in general and, more specifically, in their application to the field of vehicle reconstruction.

\tit{Acknowledgements} 
The work is partially supported by the Department of Engineering Enzo Ferrari, under the project FAR-Dip-DIEF 2022 “AI platform with digital twins of interacting robots and people”.
%
\bibliographystyle{splncs04}
\bibliography{biblio}

\end{document}